\documentclass{article} % For LaTeX2e
\usepackage{longcat_style,times}

\usepackage{amsmath,amsfonts,bm}

\def\eqref#1{equation~\ref{#1}}
\def\1{\bm{1}}

\DeclareMathAlphabet{\mathsfit}{\encodingdefault}{\sfdefault}{m}{sl}
\SetMathAlphabet{\mathsfit}{bold}{\encodingdefault}{\sfdefault}{bx}{n}

\usepackage{hyperref}
\usepackage{url}

\usepackage{graphicx}
\usepackage{balance}  % for  \balance command ON LAST PAGE  (only there!)
\usepackage{amsmath}
\usepackage{algpseudocode}
\usepackage{latexsym}
\usepackage{multirow}
\usepackage{multicol}
\usepackage{color}
\usepackage{dashrule}
\usepackage{extarrows}
\usepackage{dsfont}
\usepackage{centernot}
\usepackage{hyperref}
\usepackage{natbib}
\usepackage{fancybox}
\usepackage[utf8]{inputenc} % allow utf-8 input
\usepackage[T1]{fontenc}    % use 8-bit T1 fonts
\usepackage{hyperref}       % hyperlinks
\usepackage{url}            % simple URL typesetting
\usepackage{booktabs}       % professional-quality tables
\usepackage{amsfonts}       % blackboard math symbols
\usepackage{amsmath}
\usepackage{nicefrac}       % compact symbols for 1/2, etc.
\usepackage{microtype}      % microtypography
\usepackage{xcolor} % colors
\usepackage{amsmath}
\usepackage{lipsum}
\usepackage{colortbl}
\usepackage{subcaption}
\usepackage{tabularx}
\usepackage{makecell}
\usepackage{wrapfig}
\usepackage{bm}
\usepackage{multirow}
\usepackage{xcolor} % Load the color package
\usepackage{bbm}
\usepackage{mdframed}
\usepackage{enumitem}
\usepackage{xspace}
\usepackage{adjustbox}

\usepackage{lineno}

\usepackage{xurl}
\usepackage{subcaption}
\usepackage{amsmath}
\usepackage{multirow}
\usepackage{makecell}
\usepackage{booktabs}
\usepackage{bm}

\usepackage[T1]{fontenc}
\usepackage[utf8]{inputenc}

\usepackage{microtype}

\usepackage{inconsolata}

\usepackage{graphicx}

\usepackage{amsmath}
\usepackage{amssymb}
\usepackage{booktabs}
\usepackage{multirow}
\usepackage{makecell}
\usepackage{tcolorbox}
\usepackage{listings}
\usepackage{xcolor}
\usepackage[table]{xcolor} 
\usepackage{graphicx}
\usepackage{amsmath}

\definecolor{promptbg}{RGB}{248,248,248}
\definecolor{promptframe}{RGB}{210,210,210}

\lstdefinestyle{promptstyle}{
  basicstyle=\ttfamily\small,
  columns=fullflexible,
  breaklines=true,
  breakatwhitespace=true,
  keepspaces=true,
  showstringspaces=false,
  frame=none,
  tabsize=2,
  gobble=0,
  breakindent=0pt,         % 关键：续行不缩进
  breakautoindent=false,   % 关键：不要自动对齐缩进
  postbreak=\mbox{}        % 关键：断行后不插入任何东西
}

\newtcolorbox{promptbox}[1][Prompt]{
  colback=promptbg,
  colframe=promptframe,
  title=\textbf{#1},
  boxrule=0.6pt,
  arc=1mm,
  coltitle=black,        % 1. 标题变黑
  colbacktitle=gray!25,  % 2. 标题背景变浅灰 (配合黑色文字，否则看不清)
  boxsep=4pt,            % 3. 减小内部基准间距
  left=4pt, right=4pt, top=4pt, bottom=4pt % 4. 紧凑边距 (数值越小越紧)
}
\definecolor{darkblue}{rgb}{0, 0, 0.5}
\hypersetup{colorlinks=true, citecolor=darkblue, linkcolor=darkblue, urlcolor=darkblue}

\usepackage{listings}
\usepackage{xcolor}
\usepackage[ruled,vlined]{algorithm2e}
\usepackage{caption}
\usepackage{float} % for H option in algorithm
\usepackage{arydshln}
\usepackage{wrapfig} 
\usepackage{tcolorbox}
\tcbuselibrary{theorems}
\tcbuselibrary{most}
\usepackage[dvipsnames]{xcolor}

\definecolor{customTeal}{RGB}{0, 128, 128} 
\definecolor{emphasisColor}{RGB}{255, 0, 0} % Red color for emphasis
\definecolor{oursBlue}{RGB}{51,202,246}

\definecolor{blue1}{HTML}{508AB2}
\definecolor{green2}{HTML}{BFF6BA}

\newtcbtheorem[]{prompt}{Prompt}%
{colback=SeaGreen!10!CornflowerBlue!10,
 colframe=RoyalPurple!55!Aquamarine!100!,
 fonttitle=\bfseries,
 left=.02in, right=.02in, bottom=.02in, top=.02in,
 before upper={\linespread{1.5}\selectfont}}{prompt}

\renewcommand{\headeright}{\raisebox{-0.2\height}{\includegraphics[height=2.2em]{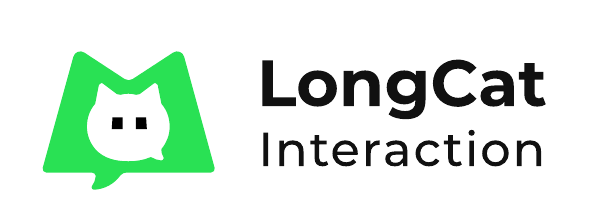}}}
\renewcommand{\shorttitle}{Signal-Calibrated On-Policy Distillation Enhancement with Dual-Path Adaptive Weighting}
\definecolor{darkblue}{rgb}{0, 0, 0.5}
\hypersetup{colorlinks=true, citecolor=darkblue, linkcolor=darkblue, urlcolor=darkblue}

\makeatletter
\renewcommand{\@maketitle}{%
  \vbox{%
    \hsize\textwidth
    \linewidth\hsize
    \vskip -0.5in
    \noindent
    \begin{minipage}{0.2\textwidth}
      \includegraphics[width=\linewidth]{pic/Logo.pdf}
    \end{minipage}%
    \\
    \rule{\linewidth}{1pt}
    \hspace{0.05\textwidth}%
    \begin{minipage}{0.8\textwidth}
    \end{minipage}

    \centering
    {\LARGE \bfseries\@title\par}
    \vskip 0.1in  % 调整这个值：0.3in=小, 0.5in=中, 0.7in=大
    \def\And{%
      \end{tabular}\hfil\linebreak[0]\hfil%
      \begin{tabular}[t]{c}\bf\rule{\z@}{24\p@}\ignorespaces%
    }
    \def\AND{%
      \end{tabular}\hfil\linebreak[4]\hfil%
      \begin{tabular}[t]{c}\bf\rule{\z@}{24\p@}\ignorespaces%
    }
    \begin{tabular}[t]{c}\bf\rule{\z@}{24\p@}\@author\end{tabular}%
  \vskip 0.05in 
  }
}
\makeatother

\title{SCOPE: Signal-Calibrated On-Policy Distillation Enhancement with Dual-Path Adaptive Weighting}

\makeatletter
\def\@fnsymbol#1{\ensuremath{\ifcase#1\or \dagger\or \ddagger\or
   \mathsection\or \mathparagraph\or \|\or **\or \dagger\dagger
   \or \ddagger\ddagger \else\@ctrerr\fi}}
\makeatother
\renewcommand{\thefootnote}{\fnsymbol{footnote}}

\author{
\begin{tabular}{c}
\textbf{Binbin Zheng}$^{1,2\ast\dagger}$
\quad
\textbf{Xing Ma}$^{2\ast\ddagger}$
\quad
\textbf{Yiheng Liang}$^{2,3}$
\quad
\textbf{Jingqing Ruan}$^{2}$ \\[1ex]
\textbf{Xiaoliang Fu}$^{4}$
\quad
\textbf{Kepeng Lin}$^{5}$
\quad
\textbf{Benchang Zhu}$^{2\ddagger}$
\quad
\textbf{Ke Zeng}$^{2}$
\quad
\textbf{Xunliang Cai}$^{2}$ \\[1ex]
\quad
\normalfont $^1$University of Science and Technology of China
\quad
$^2$Meituan LongCat Interaction Team \\[1ex]
\normalfont $^3$Nanjing University
\quad
\normalfont $^4$Fudan University
\quad
$^5$Huazhong University of Science and Technology \\[1ex]
\normalfont \texttt{\url{https://github.com/machine981/SCOPE}}
\end{tabular}
}

\begin{document}

\maketitle

% ======== 绕过模板映射，强制直接输出指定符号 ========
\begingroup
\renewcommand{\thefootnote}{$\ast$}
\footnotetext{Equal contribution.}
\renewcommand{\thefootnote}{$\dagger$}
\footnotetext{This work was done during an internship at Meituan.}
\renewcommand{\thefootnote}{$\ddagger$}
\footnotetext{Corresponding author.}
\endgroup
% ====================================================

\begin{abstract}
On-policy reinforcement learning has become the dominant paradigm for reasoning alignment in large language models, yet its sparse, outcome-level rewards make token-level credit assignment notoriously difficult. 
On-Policy Distillation (OPD) alleviates this by introducing dense, token-level KL supervision from a teacher model, but typically applies this supervision uniformly across all rollouts, ignoring fundamental differences in signal quality. 
We propose \textbf{S}ignal-\textbf{C}alibrated \textbf{O}n-\textbf{P}olicy Distillation \textbf{E}nhancement (\textbf{SCOPE}), a dual-path adaptive training framework that routes on-policy rollouts by correctness into two complementary supervision paths. For incorrect trajectories, SCOPE performs teacher-perplexity-weighted KL distillation to prioritize instances where the teacher demonstrates genuine corrective capability, while down-weighting unreliable guidance. 
For correct trajectories, it applies student-perplexity-weighted maximum likelihood estimation to concentrate reinforcement on low-confidence samples at the capability boundary rather than over-reinforcing already mastered ones. 
Both paths employ a group-level normalization to adaptively calibrate weight distributions, accounting for the intrinsic difficulty variance across prompts. 
Extensive experiments on six mathematical reasoning benchmarks show that SCOPE achieves average relative improvements of 11.42\% in Avg@32 and 7.30\% in Pass@32 over competitive baselines, with extended experiments demonstrating its broader applicability.
\end{abstract}

\section{Introduction}

In the reasoning alignment of large language models (LLMs), on-policy reinforcement learning has become the dominant paradigm, where the model samples rollouts and updates its policy based on outcome correctness~\citep{guo2025deepseek, shao2024deepseekmath, yu2025dapo}.
However, the sparse, outcome-level nature of these rewards makes token-level credit assignment notoriously difficult, often demanding massive iterations to converge~\citep{peng2026hiper, wei2025reinforcing}. On-Policy Distillation (OPD) alleviates this by introducing dense, token-level KL supervision from a teacher model on the student's self-sampled rollouts~\citep{min2024imitate, fu2026revisiting}, striking a balance between distribution consistency and training efficiency.
   
Despite its effectiveness, OPD assumes the teacher's dense supervision is \textbf{uniformly reliable} across rollouts~\citep{ko2024distillm,agarwal2024policy,fu2026revisiting}, problematic in two respects.
\textbf{(1) For incorrect trajectories}, low teacher perplexity (PPL) signifies a strong reasoning grasp, enabling reliable post-error guidance. Conversely, high PPL indicates unfamiliarity, rendering the teacher's token-level distribution an unreliable signal. 
Distillation must therefore prioritize teacher-confident instances. As shown in \S\ref{sec:preliminary}, low teacher PPL strongly correlates with successful error recovery~\citep{xiong2024can,kadavath2022language}, validating this perplexity as a proxy for genuine corrective capability.
\textbf{(2) For correct trajectories}, teacher KL supervision risks suppressing valid yet unconventional reasoning paths where the student diverges~\citep{agarwal2024policy}. While Maximum Likelihood Estimation (MLE) self-reinforcement is a natural alternative, equal-weight MLE disproportionately reinforces stably mastered samples~\citep{zhu2025surprising}, marginalizing low-confidence instances at the capability boundary. Correct trajectories should thus be weighted adaptively by the student's perplexity to maximize learning value.

The above analysis reveals a shared structural flaw in standard OPD: the absence of \textbf{signal quality awareness}. 
For incorrect trajectories, OPD fails to distinguish reliable teacher guidance from unreliable supervision.
For the correct ones, it treats all samples as equally valuable regardless of learning utility. 
Notably, the two paths require complementary weighting perspectives, with teacher perplexity applied to the former and student perplexity to the latter, motivating a unified framework that \textbf{routes trajectories by correctness} and applies \textbf{adaptive weighting tailored to each scenario}.

To this end, we propose \textbf{S}ignal-\textbf{C}alibrated \textbf{O}n-\textbf{P}olicy Distillation \textbf{E}nhancement (SCOPE), a dual-path adaptive training framework. Specifically, SCOPE routes on-policy rollouts by correctness into two supervision paths. For incorrect trajectories, it performs selective KL distillation weighted by teacher perplexity, up-weighting instances where the teacher demonstrates genuine corrective capability. For correct trajectories, it applies weighted MLE based on the student's perplexity, concentrating reinforcement on samples at the capability boundary. Finally, both paths employ a normalization mechanism that adaptively calibrates the weight distribution within each group.

Our main contributions are as follows:
\begin{itemize}[leftmargin=12pt, itemsep=1pt, parsep=0pt, topsep=0pt]
\item \textbf{Empirical analysis of signal quality heterogeneity in OPD.}
We uncover an overlooked quality variance in OPD: teacher and student perplexity reliably predict corrective capability on incorrect trajectories and capability-boundary samples on correct ones, respectively.
% (\S\ref{sec:preliminary}).
\item \textbf{The dual-path adaptive framework.} By routing rollouts based on correctness, SCOPE directs incorrect trajectories to teacher-perplexity-weighted OPD and correct trajectories to student-perplexity-weighted MLE, achieving quality-aware supervision within a unified objective.
\item \textbf{Extensive experimental validation.} SCOPE achieves an average relative improvement of 11.42\% in Avg@32 and 7.30\% in Pass@32 over competitive baselines on mathematical benchmarks, with extended experiments demonstrating its broader applicability.
\end{itemize}

\begin{figure}[t]
    \centering
    % 关键：路径和你之前一样 → figure/文件夹
    \includegraphics[width=\textwidth]{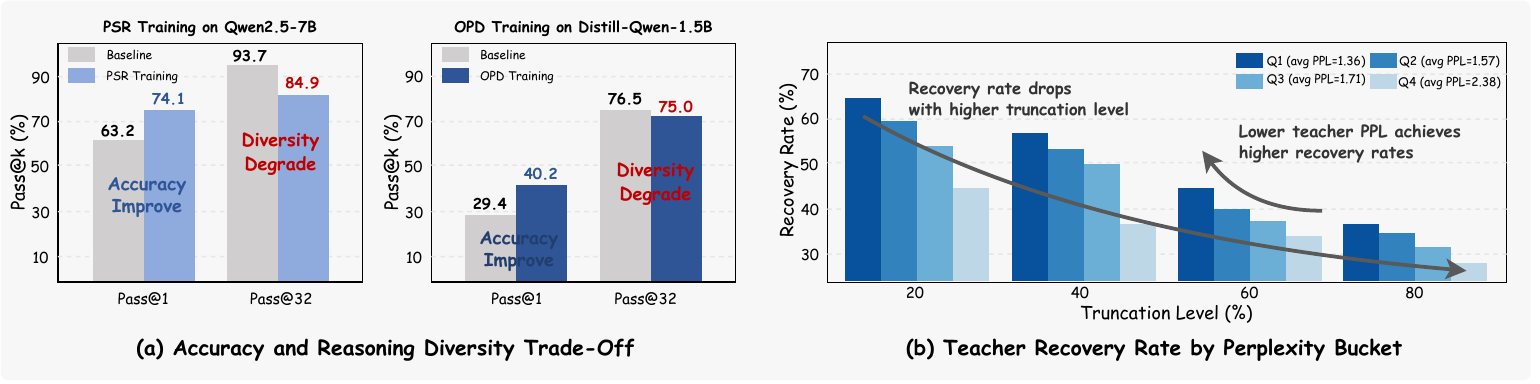}
    \caption{(a) Performance changes on the AIME24 benchmark before and after training. Both PSR and OPD training enhance pass@1 at the expense of pass@32, highlighting a clear trade-off between accuracy and reasoning diversity. (b) Recovery rate of the teacher model across varying truncation levels, conditioned on truncated student error trajectories as prefixes and stratified by perplexity.}
    \label{fig:diversity_tradeoff}
\end{figure}

\section{Preliminary Analysis}
\label{sec:preliminary}

Before presenting our framework, we conduct two empirical studies that reveal fundamental limitations of existing on-policy optimization paradigms: the degradation of reasoning diversity when optimizing successful trajectories, and the inefficiency of rectifying failed ones. These findings directly motivate the dual-path design of SCOPE.

\subsection{Diversity Degradation}
\label{subsec:diversity_loss}

Uniformly reinforcing the student's self-generated correct trajectories amplifies its dominant reasoning paths, marginalizing valid but low-probability alternatives~\citep{zhu2025surprising,li2025choice,liang2025beyond}. Meanwhile, imposing dense teacher signals forces it to match the teacher's distribution strictly, thereby suppressing the student's own valid and diverse explorations~\citep{yuan2025more}. Ultimately, both paradigms inevitably lead to severe mode collapse.

\paragraph{The Pass@$k$ Paradox.} \citet{zhu2025surprising} report that uniformly reinforcing a model's own correct answers (Positive Sample Reinforcement, PSR) on Qwen2.5-7B yields a stark paradox: Pass@1 improves, yet Pass@32 severely degrades from 93.7\% to 84.9\%. 
To investigate whether dense teacher supervision can circumvent this issue, we apply OPD to all generated trajectories of DeepSeek-R1-Distill-Qwen-1.5B, and observe a similarly striking pattern: Pass@1 increases, but Pass@32 drops from 76.5\% to 75.0\% (Figure~\ref{fig:diversity_tradeoff}a). Both results confirm that uniform optimization of correct trajectories inevitably sharpens the policy toward dominant reasoning modes at the expense of diversity.
The underlying cause is intuitive: among correct rollouts for the same prompt, some follow the dominant reasoning mode while others arrive at the answer through rare, unconventional paths. Uniform optimization treats all trajectories equally, over-reinforcing the former and extinguishing the latter. This calls for \textbf{a weighting mechanism that distinguishes well-mastered solutions from under-explored ones, allocating greater importance to the latter to preserve reasoning diversity. }

\subsection{Rectification Inefficiency}
\label{subsec:inefficient_rectification}

When a model generates incorrect trajectories, relying solely on self-exploration to find the correct path is highly inefficient in complex reasoning tasks~\citep{lu2025onpolicydistillation, zhao2026self}. Although introducing a teacher for corrective supervision is intuitive, the on-policy nature creates a severe bottleneck: the teacher risks being conditioned on flawed prefixes generated by the weaker student. If such prefixes are logically corrupted, the teacher's guidance may degenerate into noise, making rectification inefficient and unstable.

\paragraph{The Flawed Prefix Trap.}\label{para:error_recover} To investigate the efficiency of external rectification on flawed on-policy student prefixes, we conduct an Error Recovery Experiment. Specifically, we sample 2,000 problems from the DeepMath dataset~\citep{he2025deepmath} and generate reasoning trajectories using the student model (Distill-R1-Deepseek-Qwen-1.5B), retaining the incorrect ones. We then compute the perplexity of these flawed trajectories using the teacher model (Skywork-OR1-7B) and stratify them into distinct buckets based on their perplexity scores. Finally, we truncate these prefixes at various length ratios and evaluate the teacher model's recovery accuracy when prompted to complete the generation (More details and case studies are provided in Appendix~\ref{appendix:preliminary} and \ref{appendix:case_study}, respectively).

As shown in Figure~\ref{fig:diversity_tradeoff}b, low-PPL prefixes (Q1) consistently yield significantly higher recovery rates than their high-PPL counterparts (Q4) across all truncation levels, outperforming them by a margin of up to \textbf{+19.4\%}. Furthermore, as the truncation level increases, recovery rates drop rapidly across all groups, with even the best-performing group declining to approximately \textbf{35\%} at an 80\% truncation ratio. This reveals a critical mechanism: high teacher PPL indicates severe context degradation, which disrupts the teacher's reasoning process and generates high-entropy noise. Learning from these degraded regions renders rectification extremely inefficient. Conversely, low PPL ensures the prefix remains structurally coherent, allowing the teacher to provide high-quality corrective signals. This motivates our belief that \textbf{efficient rectification requires down-weighting samples with high teacher PPL to filter out misleading noise}.

\section{Methodology}
\label{sec:methodology}
\begin{figure*}[t]
    \centering
    \includegraphics[width=\textwidth]{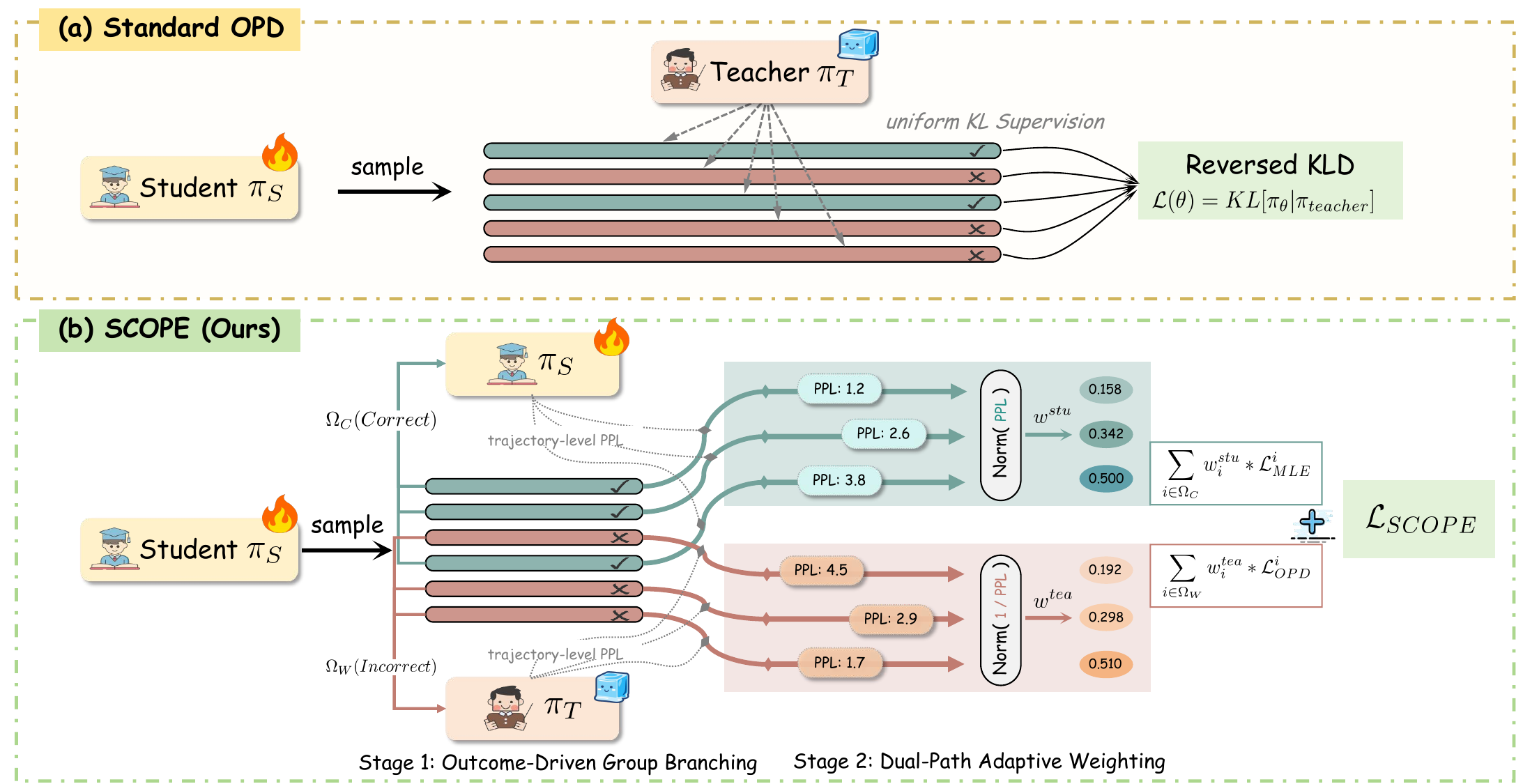}
     \caption{(a) Standard OPD applies uniform supervision to all samples. (b) Our SCOPE framework refines the learning process by first dividing trajectories into correct $\Omega_C$ and incorrect $\Omega_W$ sets, applying dual-path perplexity-based weighting, and finally optimizing the weighted branches via a unified objective.}
    \label{fig:main_pipeline}
\end{figure*}

To overcome the aforementioned degradation of reasoning diversity and the inherent inefficiency of prefix rectification, we present SCOPE, a dual-path training framework. As illustrated in Figure \ref{fig:main_pipeline}, SCOPE routes on-policy rollouts based on trajectory outcomes, and filters out misleading teacher noise through perplexity-calibrated adaptive weighting. We first describe the outcome-driven group branching (\S\ref{subsec:branching}), then detail the dual-path adaptive weighting mechanism (\S\ref{subsec:weighting}), and finally formulate the overall objective (\S\ref{subsec:objective}).

\subsection{Outcome-Driven Group Branching}
\label{subsec:branching}

During the on-policy rollout, for each input prompt $x$, the student model generates a group of $N$ responses, denoted as $Y^x = \{y_1, y_2, \dots, y_N\}$, where each response is a sequence of tokens $y_i = (a_{i,1}, a_{i,2}, \dots, a_{i,|y_i|})$. Each response $y_i \in Y^x$ is subsequently evaluated by a verifier to yield a binary reward $R_i \in \{0, 1\}$. We utilize these binary rewards as a routing signal to explicitly partition the generated trajectories into two disjoint subsets: the correct set $\Omega_c^x = \{y_i \in Y^x \,|\, R_i = 1\}$ and the incorrect set $\Omega_w^x = \{y_i \in Y^x \,|\, R_i = 0\}$.

\paragraph{On-Policy Surrogate Formulation.}
To optimize the current policy $\pi_\theta$ using trajectories generated by the behavior policy $\pi_{\text{old}}$, we account for the distribution shift by defining the token-level importance sampling ratio:
\begin{equation}
    \rho_{i,t}(\theta) = \frac{\pi_\theta(a_{i,t} \,|\, x, a_{i,<t})}{\pi_{\text{old}}(a_{i,t} \,|\, x, a_{i,<t})}
\end{equation}
Building upon this, we design two distinct surrogate objectives for the partitioned subsets:

% \begin{itemize}[leftmargin=*]
\textbf{Valid Trajectory Exploitation ($i \in \Omega_c^x$):} Correct trajectories encapsulate direct, valid reasoning steps. Rather than relying on teacher guidance, we explicitly leverage these self-generated successful attempts. By maximizing their likelihood through the surrogate objective, we reinforce the model's intrinsic capabilities:
\begin{equation}
    \label{eq:sft_loss}
    \mathcal{L}_{\text{MLE}}(x, y_i; \theta) = - \sum_{t=1}^{|y_i|} \rho_{i,t}(\theta)
\end{equation}

\textbf{Flawed Trajectory Rectification ($i \in \Omega_w^x$):} Incorrect trajectories lack inherent supervisory targets. To enable effective rectification, we leverage the teacher policy $\pi_T$ to provide external guidance. The on-policy distillation objective minimizes the reverse KL divergence by treating the token-level log-ratio as a negative advantage, yielding the following surrogate loss:
\begin{equation}
\label{eq:opd_loss}
\begin{split}
     \mathcal{L}_{\text{OPD}}(x, y_i; \theta) = \sum_{t=1}^{|y_i|} \rho_{i,t}(\theta)
    \Big( \log \pi_{\bar{\theta}}(a_{i,t} \,|\, x, a_{i,<t}) - \log \pi_T(a_{i,t} \,|\, x, a_{i,<t}) \Big),
\end{split}
\end{equation}
where $\bar{\theta}$ denotes parameters detached from the computational graph.

\subsection{Dual-Path Adaptive Weighting}
\label{subsec:weighting}

To mitigate diversity degradation and improve rectification efficiency, we introduce a novel mechanism termed \textbf{Dual-Path Adaptive Weighting (DPAW)}, which operates strictly within the candidate response group of each prompt. Let $\log \pi(y_i \,|\, x) = \sum_{t=1}^{|y_i|} \log \pi(a_{i,t} \,|\, x, a_{i,<t})$ denote the sequence-level log-probability. To intuitively quantify the trajectory's intrinsic uncertainty, we formulate our weighting mechanism using sequence perplexity, where $\text{PPL}(y_i|x) = \exp (-\frac{1}{|y_i|} \log \pi(y_i \,|\, x))$.

\paragraph{Student-guided Weight: Amplifying ``Unconventional Valid Paths''.} 
For correct trajectories ($i \in \Omega_c^x$), we want the student model to focus on instances where it successfully reaches the correct outcome through low-probability, alternative routes. To assign higher weights to these low-confidence trajectories, we apply a group-relative softmax over the length-normalized \textbf{negative} log-probabilities. Using the sequence probability $\pi_S(y_i \,|\, x)$, the \textbf{student-guided weight} $w_i^{\text{stu}}$ is computed as:
\begin{equation}
   \label{eq:weight_correct}
   \begin{aligned}
       w_i^{\text{stu}} = \frac{\exp \left( - \frac{1}{\tau |y_i|} \log \pi_S(y_i \,|\, x) \right)}{\sum_{j \in \Omega_c^x} \exp \left( - \frac{1}{\tau |y_j|} \log \pi_S(y_j \,|\, x) \right)} 
        = \frac{\text{PPL}_S(y_i \,|\, x)^{1/\tau}}{\sum_{j \in \Omega_c^x} \text{PPL}_S(y_j \,|\, x)^{1/\tau}}, \quad \forall i \in \Omega_c^x.
   \end{aligned}
\end{equation}
As shown in the rightmost term, this formulation can be elegantly expressed as a direct group-level normalization of the student's perplexity (scaled by a temperature $\tau$). Thus, correct but high-perplexity samples naturally receive amplified supervision.
A theoretical analysis is provided in Appendix \ref{appendix:explore}.

\paragraph{Teacher-guided Weight: Filtering Out ``Prefix-Induced Noise''.} 
For incorrect trajectories ($i \in \Omega_w^x$), conditioning the teacher on flawed prefixes often leads to high-entropy noise (as demonstrated in Section \ref{para:error_recover}). To prevent the student from inheriting this noise, we rely on the teacher only when it provides highly confident corrections. Specifically, we apply the softmax directly over the teacher's length-normalized log-probabilities:
\begin{equation}
   \label{eq:weight_wrong}
   % \small
   % \resizebox{0.85\columnwidth}{!}{$\displaystyle
   \begin{aligned}
       w_i^{\text{tea}} = \frac{\exp \left( \frac{1}{\tau |y_i|} \log \pi_T(y_i \,|\, x) \right)}{\sum_{j \in \Omega_w^x} \exp \left( \frac{1}{\tau |y_j|} \log \pi_T(y_j \,|\, x) \right)} 
       = \frac{\text{PPL}_T(y_i \,|\, x)^{-1/\tau}}{\sum_{j \in \Omega_w^x} \text{PPL}_T(y_j \,|\, x)^{-1/\tau}}, \quad \forall i \in \Omega_w^x.
   \end{aligned}
   % $}
\end{equation}
This formulation selectively down-weights instances where the teacher exhibits high perplexity, thereby effectively filtering out the noise induced by flawed prefixes. We formalize this filtering criterion in Appendix \ref{appendix:ppl}.

\subsection{The Overall SCOPE Objective}
\label{subsec:objective}

Finally, we integrate the outcome-driven branches and the adaptive weights into an overall objective over the dataset $\mathcal{D}$. The overall SCOPE loss $\mathcal{J}_{\text{SCOPE}}$ is formulated as:
\begin{equation}
\label{eq:prompt_loss}
\begin{split}
    \mathcal{J}_{\text{SCOPE}} = \mathbb{E}_{x \sim \mathcal{D}} \Bigg[ \sum_{i \in \Omega_c^x} w_i^{\text{stu}} \cdot \mathcal{L}_{\text{MLE}}(x, y_i)
    + \sum_{i \in \Omega_w^x} w_i^{\text{tea}} \cdot \mathcal{L}_{\text{OPD}}(x, y_i) \Bigg].
\end{split}
\end{equation}
The group-level normalization stabilizes adaptive updates across prompts of varying difficulty. A formal analysis is provided in Appendix \ref{appendix:norm}. 
Within this framework, SCOPE adaptively calibrates supervision signals at the group level: it reinforces the student's boundary capabilities on valid paths, while distilling only informative corrections from the teacher on flawed ones.

\section{Experiment}

\subsection{Experimental Setup}

\paragraph{Training Settings and Baselines.}
We employ DeepSeek-R1-Distill-Qwen-1.5B \citep{guo2025deepseek} and 
Qwen3-1.7B-Base \citep{yang2025qwen3} as student models, paired with 
SkyWork-OR1-7B \citep{he2025skywork} and Qwen3-8B-Instruct \citep{yang2025qwen3} 
as their respective teachers, all trained on DeepMath \citep{he2025deepmath}. 
We compare SCOPE against the following baselines:
\begin{itemize}[leftmargin=*, itemsep=4pt, topsep=4pt, parsep=0pt]
    \item \textbf{Group Relative Policy Optimization (GRPO)} \citep{shao2024deepseekmath}: Optimizes the policy 
    via group-relative advantages based on outcome rewards across 
    multiple sampled responses.
    \item \textbf{Knowledge Distillation (KD)} \citep{kim2016sequence}: Trains the student on static 
    teacher-generated sequences from an offline dataset via supervised 
    learning.
    \item \textbf{On-Policy Distillation (OPD)} \citep{lu2025onpolicydistillation}: Applies dense, token-level KL divergence supervision from the teacher on 
    student-sampled trajectories.
\end{itemize}

\paragraph{Evaluation Benchmarks and Metrics.}
To comprehensively assess the reasoning capabilities of our model, we measure its performance across a wide range of datasets, including MATH500 \citep{hendrycks2021measuring}, AIME24 \citep{maa2024aime}, AIME25 \citep{maa2025aime}, AMC~2023 \citep{maa2023amc}, Minerva \citep{lewkowycz2022solving}, and OlympiadBench \citep{he2024olympiadbench}. Our evaluation employs two key metrics: Avg@32, which reflects the model's expected stability, and Pass@32, which highlights its upper-bound capability.

\paragraph{Implementation Details.}
During training, we employ a global batch size of 256, a maximum prompt length of 4,096 tokens, a completion length of 12,288 tokens, a rollout temperature of $0.6$, and a weight temperature of $1.0$. For evaluation, we report performance based on a rollout temperature of $0.6$, top-$p$ sampling with $p = 0.95$, and a maximum response length of 32,768 tokens. More details are provided in Appendix~\ref{Appendix:Implementation}.

\subsection{Main Results}
\paragraph{Performance on Mathematical Reasoning.}
Table~\ref{tab:main_results} illustrates the evaluation results across challenging mathematical reasoning benchmarks. Under the primary configuration (Teacher: SkyWork-OR1-7B, Student: Distill-Qwen-1.5B), SCOPE consistently achieves the best Avg@32 performance. Compared to strong baselines, SCOPE yields an average relative improvement of +5.54\% over standard OPD, with notable gains of +10.69\% on Olympiad and +6.59\% on AMC23. These gains stem from our teacher-guided weighting, which adaptively penalizes high-perplexity failed trajectories to bypass the ``flawed prefix trap'' and thereby extract precise corrective signals. Furthermore, Pass@32 results demonstrate SCOPE's unique capability to preserve reasoning diversity and overcome the Pass@$k$ paradox, a challenge that is especially severe when optimizing raw base models. As shown in the bottom half of Table~\ref{tab:main_results}, experiments on the Qwen3-1.7B-Base reveal that standard paradigms (e.g., GRPO, KD) drastically degrade the base model's inherent exploration ability, indicating severe mode collapse. In contrast, SCOPE effectively prevents this degradation and significantly elevates the multi-sample pass rate. This improvement can be attributed to our student-guided weighting, which actively amplifies ``unconventional valid paths'' by assigning higher weights to correct but high-perplexity trajectories. 
% Ultimately, SCOPE successfully translates preserved exploration diversity into a higher upper bound of correct solutions across different model architectures.

\begin{table*}[t!]
\phantomsection 
\centering
\definecolor{headerblue}{RGB}{31, 73, 125}
\definecolor{rowgray}{RGB}{255, 255, 255}
\definecolor{accentorange}{RGB}{204, 102, 0}
\definecolor{deltagreen}{RGB}{0, 128, 64}
\definecolor{sectionbg}{RGB}{220, 230, 242}
\definecolor{deltared}{RGB}{204, 0, 0}
\renewcommand{\arraystretch}{1.15}
\caption{Main results on mathematical reasoning benchmarks under different teacher--student configurations. We report Avg@32 (A@32) and Pass@32 (P@32) for each benchmark. \textbf{Bold} denotes the best performance and \underline{underlined} the second-best.}
\label{tab:main_results}

% 为了防止缩放后文字过于拥挤，你可以保留 2pt 也可以删掉下面这行恢复默认，
% resizebox 会自动处理整体宽度
\setlength{\tabcolsep}{2pt} 

% --- 使用 resizebox 强制自适应页面宽度 ---
\resizebox{\textwidth}{!}{%
\begin{tabular}{lcccccccccccccc}
\toprule[1.2pt]
\textbf{Model} 
& \multicolumn{2}{c}{\textbf{AIME24}} 
& \multicolumn{2}{c}{\textbf{AIME25}} 
& \multicolumn{2}{c}{\textbf{AMC23}} 
& \multicolumn{2}{c}{\textbf{MATH500}} 
& \multicolumn{2}{c}{\textbf{Minerva}} 
& \multicolumn{2}{c}{\textbf{Olympiad}} 
& \multicolumn{2}{c}{\textbf{Average}} \\
\cmidrule(lr){2-3} \cmidrule(lr){4-5} \cmidrule(lr){6-7} 
\cmidrule(lr){8-9} \cmidrule(lr){10-11} \cmidrule(lr){12-13} \cmidrule(lr){14-15}
& \small A@32 & \small P@32 
& \small A@32 & \small P@32 
& \small A@32 & \small P@32 
& \small A@32 & \small P@32 
& \small A@32 & \small P@32 
& \small A@32 & \small P@32 
& \small A@32 & \small P@32 \\
\midrule[0.8pt]

%--- Section 1 ---
\multicolumn{15}{c}{%
  \cellcolor{sectionbg}%
  \textcolor{headerblue}{\textbf{\textit{%
    Teacher:~SkyWork-OR1-7B~$\longrightarrow$~Student:~DeepSeek-R1-Distill-Qwen-1.5B%
  }}}%
} \\

\rowcolor{white}
R1-Distill-Qwen-1.5B        
& 29.4 & \underline{76.5} & 23.9 & 46.9 & 72.7 & 94.7 & 84.6 & 97.3 & 32.3 & \textbf{55.9} & 44.2 & 67.5 & 47.9 & \underline{73.1} \\
\rowcolor{white}
~~w/ GRPO                       
& 35.5 & 68.3 & 24.5 & 45.1 & 75.1 & 95.0 & 87.0 & 96.7 & \underline{35.1} & 53.5 & 40.5 & 67.7 & 49.6 & 71.1 \\
\rowcolor{white}
~~w/ KD                         
& 26.6 & 71.4 & 22.2 & 45.6 & 69.1 & \underline{96.3} & 84.1 & 97.4 & 30.7 & 54.1 & 39.7 & 66.9 & 45.4 & 72.0 \\
\rowcolor{white}
~~w/ OPD                        
& \underline{40.2} & 75.0 & \underline{28.9} & \underline{48.5} & \underline{75.9} & 95.0 & \underline{89.0} & \underline{97.7} & 34.9 & 53.0 & \underline{44.9} & \underline{69.3} & \underline{52.3} & \underline{73.1} \\
\rowcolor{white}
~~w/ \textbf{SCOPE (Ours)}           
& \textbf{42.7} & \textbf{77.9} & \textbf{30.4} & \textbf{50.9} & \textbf{80.9} & \textbf{97.2} 
& \textbf{89.8} & \textbf{97.9} & \textbf{37.8} & \underline{55.1} & \textbf{49.7} & \textbf{70.9} 
& \textbf{55.2} & \textbf{75.0} \\
\rowcolor{deltagreen!10}
\textit{$\Delta\%$ vs.\ OPD}
& \textcolor{deltagreen}{\small +6.22\%} & \textcolor{deltagreen}{\small +3.87\%} 
& \textcolor{deltagreen}{\small +5.19\%}  & \textcolor{deltagreen}{\small +4.95\%} 
& \textcolor{deltagreen}{\small +6.59\%}  & \textcolor{deltagreen}{\small +2.32\%} 
& \textcolor{deltagreen}{\small +0.90\%}  & \textcolor{deltagreen}{\small +0.20\%} 
& \textcolor{deltagreen}{\small +8.31\%}  & \textcolor{deltagreen}{\small +3.96\%} 
& \textcolor{deltagreen}{\small +10.69\%} & \textcolor{deltagreen}{\small +2.31\%} 
& \textcolor{deltagreen}{\small +5.54\%}  & \textcolor{deltagreen}{\small +2.60\%} \\

\midrule[0.8pt]

%--- Section 2 ---
\multicolumn{15}{c}{%
  \cellcolor{sectionbg}%
  \textcolor{headerblue}{\textbf{\textit{%
    Teacher:~Qwen3-8B-Instruct~$\longrightarrow$~Student:~Qwen3-1.7B-Base%
  }}}%
} \\

\rowcolor{white}
Qwen3-1.7B-Base            
& 3.7 & 26.7 & 3.2 & 20.7 & 24.6 & 69.4 & 54.8 & 88.6 & 11.2 & 41.6 & 16.3 & 48.0 & 19.0 & 49.2 \\
\rowcolor{white}
~~w/ GRPO                       
& 7.5 & \underline{27.2} & 4.2 & 24.5 & 33.1 & 60.0 & 62.9 & 89.8 & \underline{25.3} & \underline{49.8} & \textbf{25.9} & 50.9 & 26.5 & 50.4 \\
\rowcolor{white}
~~w/ KD                         
& 4.6 & 18.9 & 5.8 & 24.8 & 29.8 & 66.1 & 50.2 & 86.2 & 16.6 & 43.2 & 13.3 & 41.4 & 20.1 & 46.8 \\
\rowcolor{white}
~~w/ OPD                        
& \underline{12.2} & \textbf{31.5} & \underline{10.6} & \underline{29.7} & \underline{43.1} & \underline{80.1} & \underline{67.9} & \underline{90.7} & 24.2 & 49.4 & \underline{25.3} & \underline{53.7} & \underline{30.6} & \underline{55.9} \\
\rowcolor{white}
~~w/ \textbf{SCOPE (Ours)}            
& \textbf{13.3} & \textbf{31.5} & \textbf{12.1} & \textbf{35.6} & \textbf{46.3} & \textbf{83.0} & \textbf{70.9} & \textbf{93.0} & \textbf{27.9} & \textbf{54.0} & 24.6 & \textbf{54.6} & \textbf{32.5} & \textbf{58.6} \\
\rowcolor{deltagreen!10}
\textit{$\Delta\%$ vs.\ OPD}
& \textcolor{deltagreen}{\small +9.02\%} & \textcolor{deltagreen}{\small +0.00\%} 
& \textcolor{deltagreen}{\small +14.15\%}  & \textcolor{deltagreen}{\small +19.87\%} 
& \textcolor{deltagreen}{\small +7.42\%}  & \textcolor{deltagreen}{\small +3.62\%} 
& \textcolor{deltagreen}{\small +4.42\%}  & \textcolor{deltagreen}{\small +2.54\%} 
& \textcolor{deltagreen}{\small +15.29\%}  & \textcolor{deltagreen}{\small +9.31\%} 
& \textcolor{deltared}{\small -2.77\%} & \textcolor{deltagreen}{\small +1.68\%} 
& \textcolor{deltagreen}{\small +6.21\%}  & \textcolor{deltagreen}{\small +4.83\%} \\

\bottomrule[1.2pt]
\end{tabular}%
} % 结束 resizebox
\end{table*}

\begin{figure*}[t]
    \centering
    % Subfigure a
    \begin{subfigure}{0.30\textwidth}
        \centering
        \includegraphics[width=\linewidth]{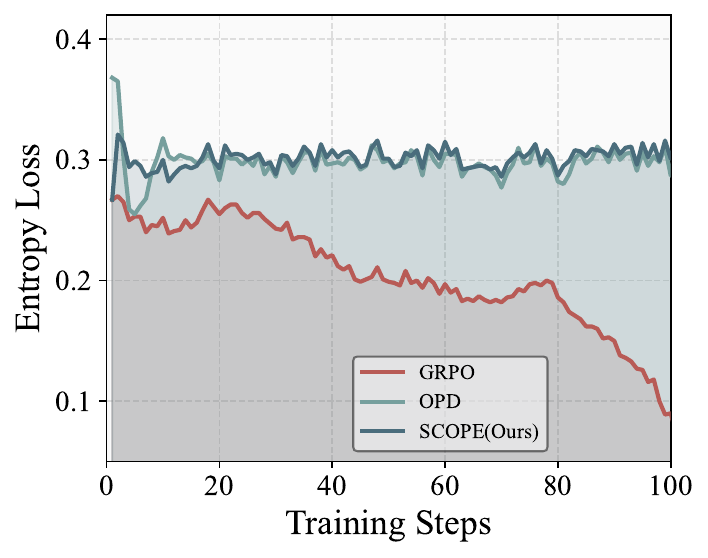}
        \caption{\footnotesize Dynamics of Entropy Loss}
        \label{fig:sub_a1}
    \end{subfigure}
    \hfill
    % Subfigure b
    \begin{subfigure}{0.30\textwidth}
        \centering
        \includegraphics[width=\linewidth]{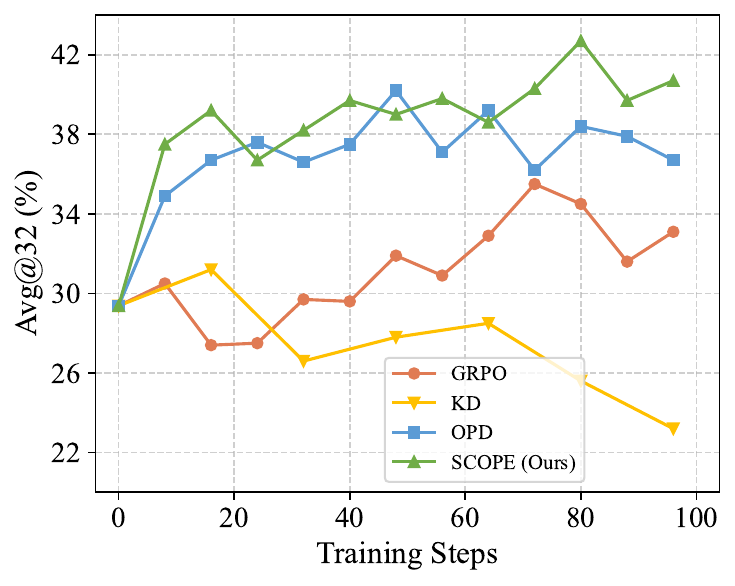} 
        \caption{\footnotesize Dynamics of AIME24 Avg@32}
        \label{fig:sub_b1}
    \end{subfigure}
    \hfill
    % Subfigure c
    \begin{subfigure}{0.30\textwidth}
        \centering
        \includegraphics[width=\linewidth]{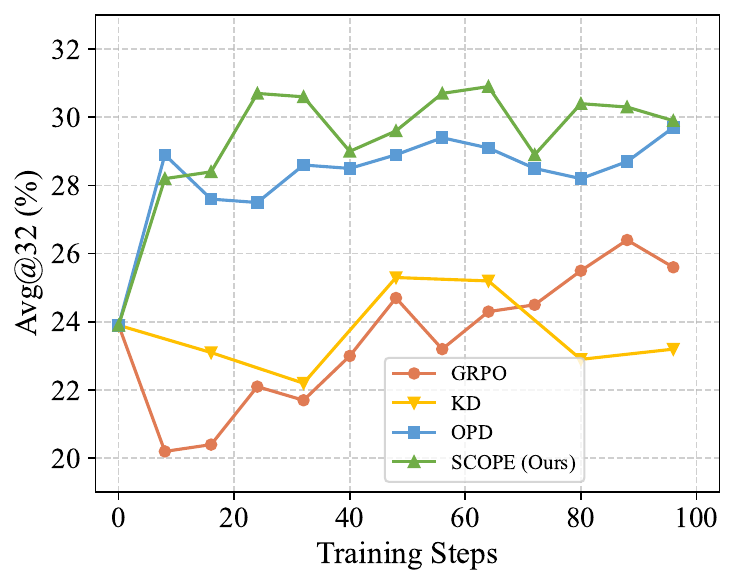}
        \caption{\footnotesize Dynamics of AIME25 Avg@32}
        \label{fig:sub_c1}
    \end{subfigure}

    \caption{Training dynamics comparing GRPO, OPD, and SCOPE (Ours): (a) entropy loss across training steps, and Avg@32 (\%) performance on (b) AIME24 and (c) AIME25.}
    \label{training_dynamics}
\end{figure*}

\paragraph{Training Dynamics.}
Figure~\ref{training_dynamics} illustrates the training dynamics of SCOPE alongside GRPO and OPD. Figure~\ref{training_dynamics}(a) shows a clear contrast in entropy evolution. While GRPO exhibits continuous entropy decay (a direct driver of the Pass@k paradox via premature exploitation), both OPD and SCOPE sustain a healthy policy entropy. However, standard OPD soon reaches a performance plateau as its uniform supervision ignores signal quality.  With dual-path adaptive weighting, SCOPE achieves consistently superior performance and sample efficiency (Figures~\ref{training_dynamics}b and c).
Overall, while GRPO underperforms due to collapsed exploration and OPD is limited by noisy distillation, SCOPE effectively breaks this bottleneck by combining diversity-preserving exploration with quality-aware error rectification.

\begin{figure*}[t!]
    \centering
    % Subfigure a
    \begin{subfigure}{0.30\textwidth}
        \centering
        \includegraphics[width=\linewidth]{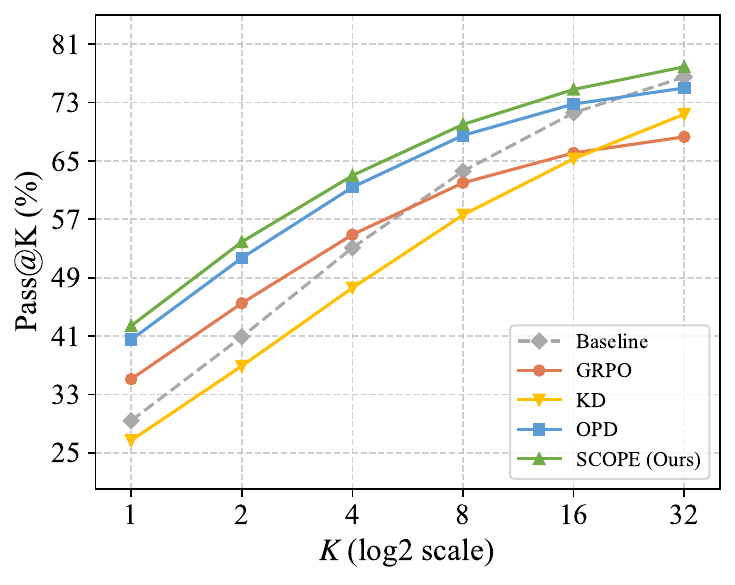} 
        \caption{\footnotesize Pass@$k$ on AIME24}
        \label{fig:sub_a2}
    \end{subfigure}%
    \hfill
    % Subfigure b
    \begin{subfigure}{0.30\textwidth}
        \centering
        \includegraphics[width=\linewidth]{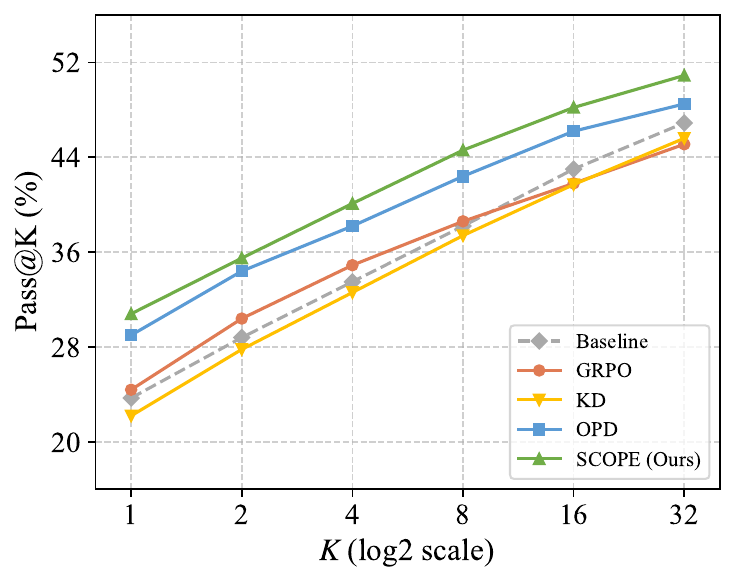}
        \caption{\footnotesize Pass@$k$ on AIME25}
        \label{fig:sub_b2}
    \end{subfigure}%
    \hfill
    % Subfigure c
    \begin{subfigure}{0.30\textwidth}
        \centering
        \includegraphics[width=\linewidth]{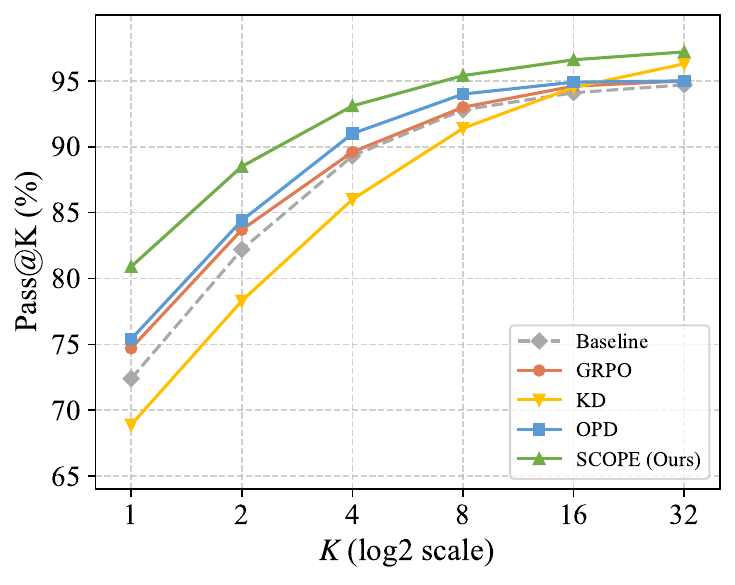}
        \caption{\footnotesize Pass@$k$ on AMC23}
        \label{fig:sub_c2}
    \end{subfigure}

    \caption{Pass@$k$ (\%) performance comparison of GRPO, OPD, and SCOPE (Ours) on the AIME24, AIME25, and AMC23 benchmarks using the DeepSeek-R1-Distill-Qwen-1.5B model.}
    \label{passk}
\end{figure*}

\begin{figure*}[t]
    \centering
    \includegraphics[width=0.98\textwidth]{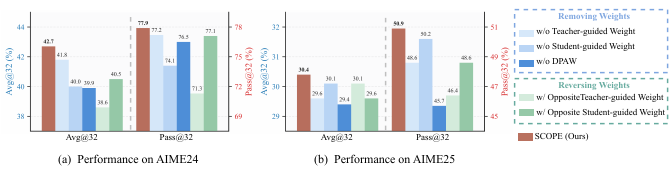}
    \caption{Ablation study on the dual-path adaptive weighting mechanism (DPAW) of SCOPE. We report the performance of Avg@32 (\%) and Pass@32 (\%) on AIME24 (a) and AIME25 (b).}
    \label{fig:ablation}
\end{figure*}

\begin{table*}[t]
\phantomsection 
\centering
\small
\definecolor{headerblue}{RGB}{31, 73, 125}
\definecolor{rowgray}{RGB}{255, 255, 255}
\definecolor{accentorange}{RGB}{204, 102, 0}
\definecolor{deltagreen}{RGB}{0, 128, 64}
\definecolor{sectionbg}{RGB}{220, 230, 242}
\definecolor{deltared}{RGB}{204, 0, 0}
\renewcommand{\arraystretch}{1.15}
\caption{Results on code generation benchmarks based on the 1.5B student model. We report Avg@32 (A@32) and Pass@32 (P@32) for each benchmark. \textbf{Bold} denotes the best performance and \underline{underlined} the second-best.}
\label{tab:code_results}
\begin{adjustbox}{width=0.75\textwidth}
\begin{tabular}{lcccccccc}
\toprule[1.2pt]
\textbf{Model} 
& \multicolumn{2}{c}{\textbf{HumanEval}} 
& \multicolumn{2}{c}{\textbf{Codeforces}} 
& \multicolumn{2}{c}{\textbf{LiveCodeBench}} 
& \multicolumn{2}{c}{\textbf{Average}} \\
\cmidrule(lr){2-3} \cmidrule(lr){4-5} \cmidrule(lr){6-7} \cmidrule(lr){8-9} 
& \small A@32 & \small P@32 
& \small A@32 & \small P@32 
& \small A@32 & \small P@32 
& \small A@32 & \small P@32 \\
\midrule[0.8pt]

%--- Section Header ---
\multicolumn{9}{c}{%
  \cellcolor{sectionbg}%
  \textcolor{headerblue}{\textbf{\textit{%
    Teacher:~SkyWork-OR1-7B~$\longrightarrow$~Student:~DeepSeek-R1-Distill-Qwen-1.5B%
  }}}%
} \\

%--- Data Rows ---
\rowcolor{white}
R1-Distill-Qwen-1.5B       
& 57.8 & 74.1 & 20.3 & 39.4 & 21.2 & 35.2 & 33.1 & 49.6 \\
\rowcolor{white}
~~w/ GRPO                     
& 61.5 & 75.2 & 32.4 & 45.3 & \textbf{34.5} & \underline{45.7} & \underline{42.8} & 55.4 \\
\rowcolor{white}
~~w/ KD                       
& 49.5 & 73.2 & 12.9 & 22.9 & 17.1 & 25.1 & 26.5 & 40.4 \\
\rowcolor{white}
~~w/ OPD                      
& \underline{64.3} & \underline{78.2} & \underline{33.3} & \underline{52.3} & 30.2 & \textbf{47.0} & 42.6 & \underline{59.2} \\
\rowcolor{white}
~~w/ \textbf{SCOPE (Ours)}         
& \textbf{67.2} & \textbf{80.2} & \textbf{34.4} & \textbf{53.7} & \underline{32.3} & \textbf{47.0} & \textbf{44.6} & \textbf{60.3} \\

%--- Delta Row ---
\rowcolor{deltagreen!10}
\textit{$\Delta\%$ vs.\ OPD}
& \textcolor{deltagreen}{\small +4.51\%} & \textcolor{deltagreen}{\small +2.56\%} 
& \textcolor{deltagreen}{\small +3.30\%} & \textcolor{deltagreen}{\small +2.68\%} 
& \textcolor{deltagreen}{\small +6.95\%} & \textcolor{deltagreen}{\small +0.00\%} 
& \textcolor{deltagreen}{\small +4.69\%} & \textcolor{deltagreen}{\small +1.86\%} \\

\bottomrule[1.2pt]
\end{tabular}
\end{adjustbox}
\end{table*}

\paragraph{Pass@$k$ Performance.} 
As illustrated in Figure~\ref{passk}, SCOPE consistently outperforms all baselines in Pass@$k$ metrics across AIME24/25 and AMC23. Notably, standard methods such as GRPO and OPD suffer from the Pass@$k$ paradox, which is particularly evident in the AIME24 evaluation. They exhibit restricted diversity scaling, where performance gains from multiple samples diminish or plateau at larger $k$. 
In contrast, SCOPE amplifies unconventional valid trajectories near the student's capability boundary, enabling the model to preserve multiple complementary reasoning routes. 
As a result, its pass rates continue to improve with larger sample sizes up to $k=32$, demonstrating stronger diversity scaling. 
These results show that SCOPE improves not only Pass@$1$ performance, but also the coverage of diverse reasoning modes, validating the effectiveness of signal-quality-aware supervision in preserving exploration diversity.

\subsection{Ablation Study}
\paragraph{Effectiveness of DPAW Mechanism.}
To validate the efficacy of the DPAW mechanism, we conduct an ablation study on the AIME24/25 benchmarks, as shown in Figure~\ref{fig:ablation}. Removing the entire DPAW module results in severe performance degradation. For example, the AIME25 Pass@32 drops significantly from 50.9\% to 45.7\%. This demonstrates that standard uniform weighting, which fundamentally ignores signal quality, fails to optimally leverage on-policy rollouts. Furthermore, omitting the student-guided weight reduces multi-sample pass rates, as AIME24 Pass@32 drops from 77.9\% to 74.1\%, while reversing its direction also hurts performance. 
Likewise, removing or reversing the teacher-guided weight compromises overall accuracy. Notably, reversing the teacher-guided weight causes a drastic drop in the AIME24 Avg@32 from 42.7\% to 38.6\%. This empirically verifies that dynamically penalizing unreliable teacher guidance effectively filters out prefix-induced noisy distillation signals. Collectively, these findings indicate that the two components of DPAW are highly complementary: the student-guided weighting maximizes exploration diversity on successful trajectories, while the teacher-guided weighting rigorously mitigates distillation noise on failed ones.

\subsection{Extended Experiments}
Table~\ref{tab:code_results} further evaluates the generality of our SCOPE on code generation tasks, including HumanEval~\citep{chen2021evaluating}, LiveCodeBench~\citep{jain2025livecodebench} and Codeforces problems. Compared with standard OPD, SCOPE consistently improves Avg@32 across all benchmarks, yielding an average relative gain of +4.69\%, and also outperforms GRPO by +4.21\% on average.  The gains over OPD are particularly notable on HumanEval and LiveCodeBench, where SCOPE increases Avg@32 by +4.51\% and +6.95\%, respectively. In addition, SCOPE improves the average Pass@32 from 59.2\% to 60.3\%, further indicating that dual-path adaptive weighting remains effective beyond mathematical reasoning. These results suggest that by distinguishing reliable teacher corrections from noisy ones while reinforcing low-confidence successful trajectories, SCOPE provides broadly applicable supervision signals for various verifiable tasks. More details can be found in Appendix~\ref{appendix:benchmarks}.

\section{Related Work}

\subsection{Reinforcement Learning with Verified Rewards}
Reinforcement learning with verified rewards (RLVR) has recently driven major advances in the reasoning capabilities of LLMs~\citep{guo2025deepseek, fu2026maspo, yu2025dapo}, leveraging deterministic outcome verifiers in objective domains (e.g., mathematics and code generation) to provide unambiguous signals that prevent reward hacking and incentivize autonomous exploration~\citep{bin2025reward, dong2025agentic}. 
However, standard RLVR algorithms such as GRPO~\citep{shao2024deepseekmath} rely on sparse, scalar outcome rewards dispensed only at the terminal step of long reasoning trajectories, severely exacerbating the credit-assignment problem~\citep{peng2026hiper, wei2025reinforcing} and depriving the model of granular process supervision~\citep{hubotter2026reinforcement}. 
This difficulty is further amplified for smaller LLMs, whose limited representational capacity leaves less room for autonomous credit propagation~\citep{xu2025kdrl, ko2026scaling}. 
While Process Reward Models (PRMs)~\citep{lightman2023let, cui2025process} can offer step-wise feedback, they demand costly human annotation and generalize poorly across domains. 
This bottleneck motivates seeking dense, token-level supervision from capable teacher models through distillation.

\subsection{Knowledge Distillation}
Knowledge distillation (KD)~\citep{hinton2015distilling} has become a primary paradigm for transferring teacher capabilities to compact student LLMs, predominantly through token-level logit alignment~\citep{gu2023minillm,agarwal2024policy, jung2025todi}.
Off-policy KD trains on static teacher-generated trajectories~\citep{guo2025deepseek,yang2025qwen3} but inherently suffers from exposure bias and distribution mismatch~\citep{agarwal2024policy, hsieh2023distilling}.
On-Policy Distillation (OPD) addresses this by optimizing student-sampled rollouts with teacher feedback via reverse KL divergence, yielding stronger convergence~\citep{ko2024distillm, lu2025onpolicydistillation}.
Recent RL-KD hybrids further unify verified rewards with teacher supervision within a single training loop: KDRL~\citep{xu2025kdrl} jointly optimizes reward and KL objectives, while RLAD~\citep{zhang2026reinforcement} and REOPOLD~\citep{ko2026scaling} inject teacher signals through dynamic reward shaping.
Despite their progress, all these methods implicitly assume that teacher supervision is \textbf{uniformly reliable} across all rollouts, overlooking the fact that teachers can be confidently wrong on specific trajectories, turning indiscriminate distillation into a vehicle for confirmation bias propagation.
This limitation calls for a \textbf{trajectory-level adaptive mechanism} that \textbf{differentiates supervision strategies} based on both \textbf{rollout correctness and signal reliability}, which is precisely the design principle behind our proposed framework.

\section{Conclusion}
In this work, we proposed \textbf{S}ignal-\textbf{C}alibrated \textbf{O}n-\textbf{P}olicy Distillation \textbf{E}nhancement (\textbf{SCOPE}), a dual-path adaptive training framework that incorporates signal quality awareness into on-policy distillation. 
SCOPE routes rollouts by correctness into two complementary supervision paths: teacher-perplexity-weighted KL distillation for incorrect trajectories to prioritize reliable corrective guidance, and student-perplexity-weighted MLE for correct trajectories to reinforce under-explored reasoning paths at the capability boundary. A unified group-level normalization adaptively calibrates weight distributions across prompts of varying difficulty.
Extensive experiments on six mathematical reasoning benchmarks show that SCOPE achieves average relative improvements of 11.42\% in Avg@32 and 7.30\% in Pass@32 over competitive baselines. 
Additional results on three code generation benchmarks further demonstrate its effectiveness and broader applicability across various tasks.

\section*{Limitation}
We acknowledge two main limitations of the current study. 
\textbf{First, SCOPE depends on automatically verifiable outcome signals to divide trajectories into correct and incorrect branches.} 
Accordingly, our experiments are conducted in domains where correctness can be reliably checked, including mathematical reasoning and code generation. 
Applying SCOPE to tasks with subjective, incomplete, or preference-driven feedback, such as open-ended dialogue or creative writing, may require additional reward models or more sophisticated verification mechanisms. 
\textbf{Second, our empirical evaluation remains constrained by computational resources. }
Although we evaluate SCOPE across multiple benchmarks and teacher--student configurations, further experiments on larger foundation models, MoE architectures, and more diverse verifiable domains remain important directions for future work.

\section*{Ethical Considerations}
We recognize the ethical responsibilities associated with research on large language models. Our experiments are conducted using publicly available datasets and benchmarks, without involving private, sensitive, or personally identifiable information. We use all datasets, benchmarks, and pretrained models under their original licenses and terms of use, and our released resources will follow the corresponding licensing requirements. We report the main training configurations, evaluation settings, and implementation details to support transparency and reproducibility. Although our method is designed to enhance model reasoning capabilities, outputs may still be unreliable in certain cases and should not be relied upon as the sole basis for high-stakes decisions. The released resources are intended for research purposes, and we encourage responsible use.

\bibliography{iclr2026_conference}
\bibliographystyle{iclr2026_conference}

\newpage
\appendix

\clearpage

\clearpage
\section{Theoretical Motivation and Derivation}
\label{sec:theoretical_motivation}

In this section, we provide a theoretical motivation and derivation for SCOPE. 
We analyze how uniform distillation on flawed prefixes may amplify noisy teacher signals, explain why student-guided weighting helps preserve low-probability but valid reasoning paths, and show how group-level normalization stabilizes adaptive updates.

\subsection{Perplexity-Aware Filtering on Flawed Prefixes}
\label{appendix:ppl}

We first analyze why uniform distillation can be detrimental on incorrect trajectories. 
In on-policy distillation, the teacher signal is evaluated on trajectories sampled by the behavior policy $\pi_{old}$, rather than under an explicit expectation over the full teacher distribution. 
Following the surrogate formulation in Section~\ref{sec:methodology}, the detached token-level log-ratio serves as an advantage-like signal:
\begin{equation}
\begin{aligned}
\mathcal{L}_{OPD}(\theta) 
    = 
    \mathbb{E}_{y \sim \pi_{old}} 
    \Bigg[
        \sum_{t=1}^{|y|} 
        \rho_t(\theta)
        \Big(
            \log \pi_{\bar{\theta}}(a_t \mid x, a_{<t})
            - \log \pi_T(a_t \mid x, a_{<t})
        \Big)
    \Bigg],
\end{aligned}
\end{equation}
where $\rho_t(\theta)=\pi_\theta(a_t \mid x,a_{<t}) / \pi_{old}(a_t \mid x,a_{<t})$, and $\bar{\theta}$ denotes parameters detached from the computational graph. Evaluating the gradient near the behavior policy, i.e., $\theta \approx \theta_{old}$, gives
\begin{equation}
\begin{aligned}
\nabla_\theta &\mathcal{L}_{OPD}(\theta)
    \approx
    \mathbb{E}_{y \sim \pi_{old}}
    \Bigg[
        \sum_{t=1}^{|y|}
        \nabla_\theta 
        \log \pi_\theta(a_t \mid x,a_{<t}) \cdot
        \Big(
            \log \pi_{\bar{\theta}}(a_t \mid x,a_{<t})
            - \log \pi_T(a_t \mid x,a_{<t})
        \Big)
    \Bigg].
\end{aligned}
\end{equation}
Let $
A_t
=
\log \pi_{\bar{\theta}}(a_t \mid x,a_{<t})
-
\log \pi_T(a_t \mid x,a_{<t})
$
denote the detached distillation signal. 
For a single sampled trajectory, the corresponding stochastic gradient estimator is
\begin{equation}
\begin{aligned}
g_{OPD}(x,y)
    =
    \sum_{t=1}^{|y|}
    \nabla_\theta
    \log \pi_\theta(a_t \mid x,a_{<t})
    \cdot A_t .
\end{aligned}
\end{equation}

When the prefix $a_{<t}$ forms a flawed reasoning context, the teacher may become less reliable on the continuation. 
It may assign lower probability to the sampled token or produce a less informative distribution, increasing $-\log \pi_T(a_t \mid x,a_{<t})$ \citep{xiong2024can}. 
As a result, $A_t$ may be dominated by low-confidence evaluations rather than meaningful corrective directions.

To see how this affects optimization, consider the squared norm of the single-trajectory gradient estimator. 
By Cauchy's inequality,
\begin{equation}
\begin{aligned}
& \left\|
    g_{OPD}(x,y)
\right\|^2
    =
    \left\|
        \sum_{t=1}^{|y|}
        \nabla_\theta
        \log \pi_\theta(a_t \mid x,a_{<t})
        \cdot A_t
    \right\|^2  \le
    |y|
    \sum_{t=1}^{|y|}
    \left\|
        \nabla_\theta
        \log \pi_\theta(a_t \mid x,a_{<t})
    \right\|^2
    \cdot A_t^2 .
\end{aligned}
\end{equation}
If the score function is bounded as
$\|\nabla_\theta \log \pi_\theta(a_t \mid x,a_{<t})\| \le G$, then
\begin{equation}
\begin{aligned}
\mathbb{E}
\left[
    \left\|
        g_{OPD}(x,y)
    \right\|^2
\right]
    &\le
    |y|G^2
    \sum_{t=1}^{|y|}
    \mathbb{E}
    \left[
        A_t^2
    \right].
\end{aligned}
\end{equation}
This bound shows that the second moment of the stochastic update is controlled by the magnitude of the detached distillation signal. 
Thus, low teacher likelihood on flawed prefixes can enlarge the update scale and introduce noisy gradients.

Trajectory-level teacher perplexity summarizes this low-confidence regime:
\begin{equation}
\begin{aligned}
\mathrm{PPL}_T(y \mid x)
    &=
    \exp
    \left(
        -\frac{1}{|y|}
        \log \pi_T(y \mid x)
    \right),
\end{aligned}
\end{equation}
where 
$\log \pi_T(y \mid x)=
\sum_{t=1}^{|y|}
\log \pi_T(a_t \mid x,a_{<t})$.
A high $\mathrm{PPL}_T(y \mid x)$ indicates that the teacher assigns low average likelihood to the sampled trajectory. 
For incorrect trajectories, this often suggests that the teacher is conditioned on a degraded prefix, making its token-level guidance less reliable.

Motivated by this interpretation, SCOPE down-weights incorrect trajectories with high teacher perplexity:
\begin{equation}
\begin{aligned}
w_i^{tea}
    &=
    \frac{
        \mathrm{PPL}_T(y_i \mid x)^{-1/\tau}
    }{
        \sum_{j \in \Omega_w^x}
        \mathrm{PPL}_T(y_j \mid x)^{-1/\tau}
    } .
\end{aligned}
\end{equation}
Therefore, incorrect trajectories with high teacher likelihood are emphasized, while high-perplexity trajectories, which are more likely to contain prefix-induced teacher noise, are suppressed. 
This explains how teacher-guided weighting improves rectification efficiency by reducing the influence of likely noisy teacher signals.

\subsection{Exploration of Valid Trajectories}
\label{appendix:explore}

We next analyze why student-guided weighting helps alleviate the Pass@$k$ paradox caused by uniform reinforcement of successful trajectories. 
For correct trajectories $\Omega_c^x$, standard MLE reinforces all successful samples uniformly:
\begin{equation}
\begin{aligned}
\nabla_\theta \mathcal{L}_{MLE}
    \propto
    -
    \sum_{i \in \Omega_c^x}
    \nabla_\theta
    \log \pi_\theta(y_i \mid x).
\end{aligned}
\end{equation}
However, under on-policy sampling, correct trajectories are not equally represented. 
Dominant reasoning paths are sampled more frequently, while rare but valid paths appear with much lower probability. 
Uniform reinforcement can therefore amplify dominant modes and weaken the contribution of low-probability valid paths.

SCOPE counteracts this effect by assigning larger weights to correct trajectories with higher student perplexity:
\begin{equation}
\begin{aligned}
w_i^{stu}
    =
    \frac{
        \mathrm{PPL}_S(y_i \mid x)^{1/\tau}
    }{
        \sum_{j \in \Omega_c^x}
        \mathrm{PPL}_S(y_j \mid x)^{1/\tau}
    } .
\end{aligned}
\end{equation}
Ignoring the within-group normalization constant and focusing on relative weighting, this behaves like an inverse-probability-style correction:
\begin{equation}
\begin{aligned}
w_i^{stu}
    \propto
    \exp
    \left(
        -\frac{1}{\tau |y_i|}
        \log \pi_S(y_i \mid x)
    \right)            =
    \pi_S(y_i \mid x)^{-\frac{1}{\tau |y_i|}} .
\end{aligned}
\end{equation}
The length normalization is retained in implementation to avoid systematically favoring shorter or longer trajectories.

At the population level, omitting length normalization only for notational clarity, the weighted gradient can be written as
\begin{equation}
\begin{aligned}
\mathbb{E}_{y \sim \pi_S}
\big[
    &w^{stu}(y)
    \nabla_\theta \log \pi_\theta(y \mid x)
\big]                                       \propto
    \sum_{y \in \Omega_c^x}
    \pi_S(y \mid x)^{1-\frac{1}{\tau |y|}}
    \cdot
    \nabla_\theta \log \pi_\theta(y \mid x).
\end{aligned}
\end{equation}
This weighting partially offsets the sampling-frequency bias of on-policy trajectories by reducing the exponent of $\pi_S(y \mid x)$ from $1$ to $1-\frac{1}{\tau |y|}$. 
A smaller $\tau$ strengthens this compensation, whereas a larger $\tau$ moves the weighting closer to the uniform case. 
Although rare paths are not guaranteed to be preserved, this weighting reduces the dominance of frequent reasoning modes over low-probability valid trajectories. 
This explains how SCOPE preserves reasoning diversity while still exploiting successful on-policy samples.

\subsection{Group-Level Normalization and Stability}
\label{appendix:norm}

Finally, we analyze the stabilizing effect of group-level normalization. 
Different prompts can exhibit different perplexity scales due to variations in difficulty, length, and reasoning structure. 
Using raw perplexity-based weights may therefore make the update scale sensitive to prompt-specific statistics.

SCOPE avoids this issue by applying local softmax normalization within each trajectory group:
\begin{equation}
\begin{aligned}
w_i
    =
    \frac{
        \exp(S_i/\tau)
    }{
        \sum_{j \in \Omega}
        \exp(S_j/\tau)
    },
\end{aligned}
\end{equation}
where $S_i$ denotes the path-specific score. 
For correct trajectories, $S_i$ corresponds to the student-perplexity score; for incorrect trajectories, it corresponds to the inverse teacher-perplexity score. 
This normalization ensures
\begin{equation}
\begin{aligned}
\sum_{i \in \Omega} w_i = 1,
\qquad
0 < w_i < 1 .
\end{aligned}
\end{equation}

For a given prompt $x$, the SCOPE gradient can be written as
\begin{equation}
\begin{aligned}
\nabla_\theta \mathcal{J}_{SCOPE}(x)
    =
    \sum_{i \in \Omega_c^x}
    w_i^{stu}
    \nabla_\theta \mathcal{L}_{MLE}(x,y_i)   +
    \sum_{i \in \Omega_w^x}
    w_i^{tea}
    \nabla_\theta \mathcal{L}_{OPD}(x,y_i).
\end{aligned}
\end{equation}
By the triangle inequality,
\begin{equation}
\begin{aligned}
\left\|
    \nabla_\theta \mathcal{J}_{SCOPE}(x)
\right\|
    \le
    \sum_{i \in \Omega_c^x}
    w_i^{stu}
    \left\|
        \nabla_\theta \mathcal{L}_{MLE}(x,y_i)
    \right\|                                  +
    \sum_{i \in \Omega_w^x}
    w_i^{tea}
    \left\|
        \nabla_\theta \mathcal{L}_{OPD}(x,y_i)
    \right\| .
\end{aligned}
\end{equation}
Since the weights are normalized within each group, each branch forms a convex combination of trajectory-level gradients rather than an unbounded rescaling by raw perplexity values. 
In particular,
\begin{equation}
\begin{aligned}
\left\|
    \sum_{i \in \Omega}
    w_i g_i
\right\|
    &\le
    \sum_{i \in \Omega}
    w_i
    \left\|
        g_i
    \right\|                                 \le
    \max_{i \in \Omega}
    \left\|
        g_i
    \right\| .
\end{aligned}
\end{equation}
Thus, group-level normalization controls the update scale while preserving the intended adaptive weighting effect. 
It does not make the estimator unbiased; instead, it introduces a signal-calibrated bias that emphasizes reliable teacher corrections on flawed trajectories and underexplored valid paths on correct trajectories.

\section{Experimental Details}
\label{Experimental_Details}

\subsection{Implementation Setup}
\label{Appendix:Implementation}
\paragraph{Training Infrastructure.} 
All experiments were conducted on a high-performance distributed cluster using a total of 20 NVIDIA A100 (80GB) GPUs. Specifically, 16 GPUs (across two nodes) were allocated for training the student model, while the remaining 4 GPUs (on a single node) were dedicated to deploying the teacher model.

\paragraph{{Hyperparameter Configuration.}}
The detailed experimental settings for our study are presented in two parts. Table \ref{tab:training_config} outlines the specific training configurations, including optimization and reinforcement learning hyperparameters, for GRPO, OPD, and SCOPE. For the evaluation phase, we adopt a consistent set of generation parameters across all models, as detailed in Table \ref{tab:evaluation_parameters}. For Qwen3-1.7B-Base, due to severe repetition issues observed during evaluation, we increased the repetition penalty to mitigate this problem.

\subsection{Evaluation Benchmarks}
\label{appendix:benchmarks}
\paragraph{Mathematical Reasoning Experiment.}
To provide a comprehensive assessment of mathematical reasoning, we evaluate our method on a broad suite of benchmarks covering various problem sources, difficulty levels, and reasoning patterns. These benchmarks range from standard high-school competition problems to challenging olympiad-style tasks, with detailed characteristics in Table~\ref{tab:benchmark_characteristics}.

\paragraph{Code Generation Experiment.}
For code generation experiments, we utilize the TACO dataset~\citep{li2023taco} as our primary training corpus, which comprises 25,202 programming problems sourced from diverse competitive programming platforms including Codeforces, AtCoder, Aizu Online Judge, and GeeksforGeeks.
For evaluation, we employ the LiveCodeBench (from 2024.08 to 2025.02) and HumanEval benchmarks. Additionally, we sample 500 Code contests problems (Codeforces) from the TACO dataset as an evaluation set. To prevent data leakage, these sampled problems are explicitly excluded from our training corpus. A detailed description of the datasets is provided in Table~\ref{tab:benchmark_characteristics}.

% 第一个表格：合并的训练参数表
\begin{table}[t]
\centering
\footnotesize
\setlength{\tabcolsep}{3pt}
\caption{Training config for GRPO, OPD, and SCOPE}
\label{tab:training_config}
\begin{tabular}{lccc}
\toprule
\textbf{Parameter} & \textbf{GRPO} & \textbf{OPD} & \textbf{SCOPE} \\
\midrule
\multicolumn{4}{c}{\textit{Optimization Parameters}} \\
\midrule
Learning Rate & $5 \times 10^{-5}$ & $5 \times 10^{-5}$ & $5 \times 10^{-5}$ \\
LR Scheduler & Constant & Constant & Constant \\
Weight Decay & 0.01 & 0.01 & 0.01 \\
Train Batch Size (Prompts) & 256 & 256 & 256 \\
Mini Batch Size (Prompts) & 256 & 256 & 256 \\
\midrule
\multicolumn{4}{c}{\textit{RL \& Generation Parameters}} \\
\midrule
Max Prompt Length & 4096 & 4096 & 4096 \\
Max Completion Length & 12288 & 12288 & 12288 \\
Number of Generations ($G$) & 8 & 1 & 8 \\
Temperature & 0.6 & 0.6 & 0.6 \\
KL Coefficient ($\beta$) & 0.0001 & -- & -- \\
Clip Ratio ($\epsilon$) & 0.2 & -- & -- \\
Update Epochs per Batch & 1 & 1 & 1 \\
\bottomrule
\end{tabular}
\end{table}

% 第二个表格：统一的评测参数表
\begin{table}[t]
\centering
\small
\caption{Evaluation parameters for all models.}
\label{tab:evaluation_parameters}
\begin{tabular}{lc}
\toprule
\textbf{Parameter} & \textbf{Value} \\
\midrule
Max Tokens & 32768 \\
Temperature & 0.6 \\  % 补充了评测时的温度
Top-p & 0.95 \\
Top-k & 20 \\
Repetition Penalty (Qwen3-1.7B) & 1.08 \\
Repetition Penalty (Distill-1.5B) & 1.0 \\
Samples per Prompt & 32 \\
\bottomrule
\end{tabular}
\end{table}

\section{Supplementary Experimental Analysis}
\subsection{Preliminary Experiment}
\label{appendix:preliminary}

We sample 2,000 problems from the DeepMath dataset and generate 4 reasoning trajectories per problem using the student model (DeepSeek-R1-Distill-Qwen-1.5B) with temperature $\tau=0.6$, top-$k=20$, top-$p=0.95$, and a maximum response length of 32,768 tokens.

For each incorrect trajectory, we compute its perplexity score under the teacher model (SkyWork-OR1-7B) over the response tokens only (excluding the prompt), defined as:
\begin{equation}
    PPL(y_{w} \,|\, x) = \exp(-\frac{1}{|y_{w}|}\sum_{t=1}^{|y_{w}|}\log P_T(y_t \,|\, x))
\end{equation}
where $y_{w}$ represents the wrong samples. They are stratified into four equal-sized buckets (Q1--Q4) based on their PPL scores via quartile splitting. Table~\ref{tab:ppl_stats} reports the PPL and negative log-likelihood (NLL) statistics for each bucket.

\begin{table*}[t]
\centering
\small
\caption{Overview of evaluation benchmarks and their key characteristics, covering mathematical reasoning and code generation tasks.}
\label{tab:benchmark_characteristics}
\begin{tabular}{p{0.15\textwidth} p{0.78\textwidth}}
\toprule
\textbf{Benchmark} & \textbf{Evaluation Focus \& Characteristics} \\
\midrule
\textbf{AIME 2024} 
& Evaluates precise arithmetic and algebraic reasoning under strict answer-format constraints, where solutions are typically given as integer-valued responses. \\

\textbf{AIME 2025} 
& Serves as a held-out benchmark for assessing whether models can handle unseen and challenging mathematical reasoning patterns. \\

\textbf{AMC 2023} 
& Includes foundational competition-style problems in algebra, geometry, and combinatorics, serving as a measure of core mathematical problem-solving ability. \\

\textbf{MATH-500} 
& Uses a representative subset of the MATH benchmark spanning multiple mathematical areas, requiring symbolic manipulation, formula interpretation, and robust answer parsing. \\

\textbf{Minerva} 
& Consists of STEM-oriented mathematical problems, emphasizing technical notation, quantitative reasoning, and multi-step solution construction. \\

\textbf{OlympiadBench} 
& Collects difficult olympiad-style problems, targeting rigorous logical reasoning, advanced mathematical generalization, and complex problem-solving skills. \\

\textbf{HumanEval} 
& Comprises 164 hand-crafted Python programming problems with unit-test-based evaluation, requiring syntactically correct code generation and functional correctness against hidden test assertions. \\

\textbf{Codeforces} 
& Sourced from competitive programming platform with 500 problems, featuring varying test coverage (1--343 I/O pairs per problem) that demands robust algorithmic reasoning and edge-case handling. \\

\textbf{LiveCodeBench} 
& Curated from recent competitive programming contests , containing 279 problems with consistent multi-case evaluation, emphasizing solution reliability under diverse input conditions. \\

\bottomrule
\end{tabular}
\end{table*}

\begin{table}[t]
\small
\centering
\caption{Teacher PPL and NLL statistics for each perplexity bucket over incorrect student trajectories.}
\label{tab:ppl_stats}
\begin{tabular}{lcccc}
    \toprule
    \textbf{Bucket} & \textbf{PPL Mean} & \textbf{NLL Mean}\\
    \midrule
    Q1 (Lowest PPL) & 1.361 & 0.305 \\
    Q2              & 1.565 & 0.448 \\
    Q3              & 1.710 & 0.536 \\
    Q4 (Highest PPL)& 2.383 & 0.688 \\
    \midrule
    Overall         & 1.755 & 0.494 \\
    \bottomrule
\end{tabular}
\end{table}

\begin{table}[t]
\centering
\small
\caption{Teacher error recovery rate (\%) under different truncation ratios and PPL buckets. Each cell reports the mean accuracy over $n=4$ completions per sample. The Q1--Q4 gap (rightmost column) quantifies the within-truncation-level spread attributable to PPL stratification.}
\label{tab:exp2_full}
{
\begin{tabular}{cccccc}
\toprule
\textbf{Trunc. Ratio} & \textbf{Q1} & \textbf{Q2} & \textbf{Q3} & \textbf{Q4} & \textbf{Q1--Q4 Gap} \\
\midrule
0.2 & 64.9 & 59.7 & 54.0 & 45.4 & +19.4 \\
0.4 & 55.8 & 53.2 & 50.1 & 40.2 & +15.6 \\
0.6 & 44.8 & 43.1 & 41.0 & 34.5 & +10.3 \\
0.8 & 35.8 & 35.3 & 32.6 & 28.6 & +7.2  \\
\bottomrule
\end{tabular}
}
\end{table}

For the prefix truncation experiment, each incorrect trajectory is truncated at the nearest newline boundary to the target truncation ratio $r \in \{0.2, 0.4, 0.6, 0.8\}$, yielding a flawed prefix $y_{\text{prefix}}$. The teacher is then prompted to complete the generation from $y_{\text{prefix}}$ using the completions API with temperature $\tau=0.6$. Each prefix is completed $n=4$ times, and the recovery rate is computed as the mean accuracy over these completions. Table~\ref{tab:exp2_full} presents the complete error recovery rates across all truncation ratios and PPL buckets.

\subsection{Impact of Weight Temperature}
\begin{figure*}[t!]
    \centering
    \includegraphics[width=0.8\textwidth]{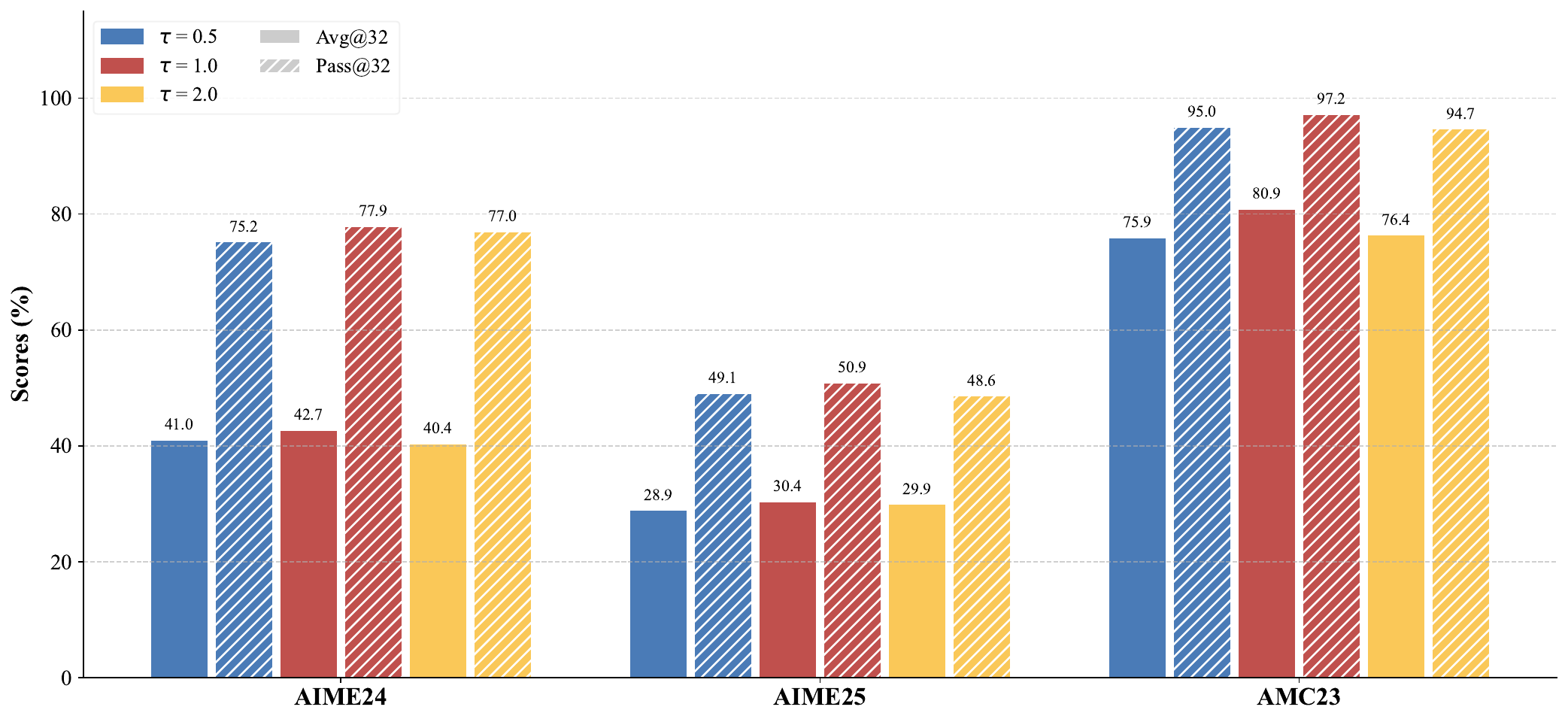}
    \caption{Impact of the temperature hyperparameter $\tau$ on model performance across AIME24, AIME25, and AMC23. The results indicate that the default configuration of $\tau = 1.0$ consistently yields the best performance across all benchmarks compared to $\tau = 0.5$ and $\tau = 2.0$.}
    \label{Sensitivity}
\end{figure*}

To examine how the sharpness of the weight distribution affects optimization, we vary the temperature parameter $\tau$ among 0.5, 1.0, and 2.0. Figure~\ref{Sensitivity} shows the effect of $\tau$ in the group-wise softmax normalization of DPAW (Eq.~\ref{eq:weight_correct} and Eq.~\ref{eq:weight_wrong}). We adopt $\tau = 1.0$ as the default setting. A smaller temperature overly sharpens the weight distribution, making the model concentrate on trajectories with extreme perplexity values. This aggressive weighting can amplify outlier noise instead of extracting reliable corrective signals, leading to unstable training. In contrast, a larger temperature flattens the distribution and weakens DPAW toward the uniform weighting used in standard OPD.  As discussed in Section~\ref{sec:preliminary}, uniform weighting lacks signal-quality awareness: it neither filters noisy teacher guidance on failed trajectories nor sufficiently emphasizes underexplored valid paths on successful ones. This can further induce the Pass@$k$ paradox and diversity collapse. Overall, $\tau = 1.0$ provides a balanced calibration of signal-quality variance, enabling effective noise filtering for incorrect paths while preserving reasoning diversity for correct ones.

\subsection{Computational Cost}
\label{sec:computational_cost}

As illustrated in Table~\ref{tab:compute_cost}, we analyze the per-step wall-clock time of SCOPE against the two primary baselines, GRPO and OPD, to characterize the additional overhead introduced by our method. All experiments are conducted on the same configuration and timing statistics are collected over the stable training region. 

While the rollout generation time is comparable to GRPO, our approach incurs an additional time overhead primarily due to teacher model queries. Notably, we use a naive synchronous training architecture where rollout and teacher logprob acquisition time do not overlap. By implementing an asynchronous  strategy, the training efficiency is expected to be comparable to that of GRPO. Furthermore, the computational overhead introduced by the weight calculation itself is minimal.

\begin{table}[t]
\centering
\small
\caption{Per-step wall-clock time breakdown (seconds) for each training method. Values are means over the stable training region. The number of rollouts is set to 8 for GRPO and SCOPE, and 1 for OPD.}
\label{tab:compute_cost}
\setlength{\tabcolsep}{6pt}
\begin{tabular}{lrrr}
\toprule
\textbf{Component} & \textbf{GRPO} & \textbf{OPD} & \textbf{SCOPE (Ours)} \\
\midrule
Generation          & 264.5 & 164.5 & 247.7 \\
Old Logprob         &  34.0 &   5.2 &  31.4 \\
Reward Computation  &   4.8 &   0.7 &   4.6 \\
Actor Update        & 151.8 &  22.9 & 154.1 \\
Teacher Scoring     &  ---  &  31.2 & 200.0 \\
\midrule
\textbf{Total (step)} & \textbf{459.0} & \textbf{227.5} & \textbf{641.9} \\
\bottomrule
\end{tabular}
\end{table}

\begin{table*}[t]
\centering
\small
\caption{Case 1: Numerical collapse.}
\label{tab:case1}
\begin{tabular}{p{2.2cm} p{10.5cm}}
\toprule
\textbf{Problem} &
Let $a_1, a_2, \ldots, a_{100}$ be integers such that
$\frac{\sum a_i^2}{\sum a_i} = 100.$
Find the maximum possible value of $a_1$. \\
\midrule
\textbf{Ground Truth} & $550$ \\
\midrule
\textbf{Student Answer} & $100$ \\
\midrule
\textbf{Error Type} & Numerical collapse \\
\midrule
\textbf{Key Excerpt} &
\begin{minipage}[t]{10.5cm}
\small
\texttt{...}$10100m^2 + 202md + 2d^2 - 2000000m - 20000d +$\\
\texttt{9999...9} \emph{[thousands of repeated digits]}\\
\texttt{$\times\, 6999 \times$ 9999...9} \emph{[continued overflow]}\\
\texttt{That seems like I'm not doing this... The problem is just to be solved.}\\
\texttt{...}$\verb|\boxed{100}|$
\end{minipage} \\
\bottomrule
\end{tabular}
\end{table*}

\subsection{Case Study}
\label{appendix:case_study}

We present representative incorrect student trajectories from the highest-PPL bucket to illustrate the qualitative nature of high perplexity errors. The cases in Tables~\ref{tab:case1}--\ref{tab:case4} illustrate the Flawed Trajectory Trap across high-perplexity ($\mathrm{PPL} \geq 1.80$) errors, encompassing both structural collapse (e.g., numerical overflow in Case 1, infinite loops in Case 2) and logical hallucinations (e.g., flawed premises in Case 3, self-contradictions in Case 4). In all such instances, the severely degraded reasoning context disrupts the teacher model, flattening its predictive distribution and forcing it to output high-entropy, uninformative noise. Standard on-policy distillation on these samples disastrously compels the student to mimic this confusion. Our Dual-Path Adaptive Weighting (DPAW) explicitly circumvents this trap. By scaling distillation weights inversely with teacher perplexity, DPAW assigns near-zero weights to these toxic trajectories. This mechanism inherently filters out context-induced hallucinations, ensuring the student only learns from structurally coherent prefixes that elicit precise corrective signals.

\begin{table*}[t]
\centering
\small
\caption{Case 2: Infinite reasoning loop.}
\label{tab:case2}
\begin{tabular}{p{2.2cm} p{10.5cm}}
\toprule
\textbf{Problem} &
Determine whether the perimeter of the Mandelbrot set is finite or infinite. \\
\midrule
\textbf{Ground Truth} & \texttt{infinite} \\
\midrule
\textbf{Student Answer} & \texttt{finite} \\
\midrule
\textbf{Error Type} & Repetitive reasoning loop; fails to commit to correct conclusion \\
\midrule
\textbf{Key Excerpt} &
\begin{minipage}[t]{10.5cm}
\small
\texttt{Okay, so I have this question... I remember that the Mandelbrot set is}\\
\texttt{this intricate fractal...}\\
\emph{[paragraph repeated verbatim 4 times]}\\
\texttt{But the question isn't about whether a point is in the set, it's about}\\
\texttt{the perimeter. I know that fractals often have infinite perimeters, but}\\
\texttt{I'm not entirely sure about the Mandelbrot set...}\\
\emph{[loop continues without resolution]}
\end{minipage} \\
\bottomrule
\end{tabular}
\end{table*}

\vspace{-10mm} 

\begin{table*}[t]
\centering
\small
\caption{Case 3: Incorrect application of a theorem.}
\label{tab:case3}
\begin{tabular}{p{2.2cm} p{10.5cm}}
\toprule
\textbf{Problem} &
Is a real-valued function $F$ on $[a,b]$ with bounded variation and the intermediate value property guaranteed to be continuous? \\
\midrule
\textbf{Ground Truth} & Yes \\
\midrule
\textbf{Student Answer} & No \\
\midrule
\textbf{Error Type} & Incorrect application of Cantor function as counterexample \\
\midrule
\textbf{Key Excerpt} &
\begin{minipage}[t]{10.5cm}
\small
\texttt{...the Cantor function is constant on the middle-thirds Cantor set,}\\
\texttt{so it's not continuous, but it's still of bounded variation.}\\
\texttt{[...]}\\
\texttt{Therefore, it's clear that such functions can be discontinuous.}\\
\texttt{So, the answer is no, it's not guaranteed to be continuous.}
\end{minipage} \\
\midrule
\textbf{Note} & The Cantor function does \emph{not} satisfy the intermediate value property, making this a flawed counterexample. The student fails to verify the premise. \\
\bottomrule
\end{tabular}
\end{table*}

\vspace{-10mm} 

\begin{table*}[t]
\centering
\small
\caption{Case 4: Off-by-one arithmetic error in integral approximation.}
\label{tab:case4}
\begin{tabular}{p{2.2cm} p{10.5cm}}
\toprule
\textbf{Problem} &
Find the integer value of $K = \left\lfloor\sum_{r=1}^{80}\int_0^1 x^{\sqrt{r}-1}\,dx\right\rfloor$. \\
\midrule
\textbf{Ground Truth} & $16$ \\
\midrule
\textbf{Student Answer} & $17$ \\
\midrule
\textbf{Error Type} & Incorrect integral bounds in Euler--Maclaurin approximation \\
\midrule
\textbf{Key Excerpt} &
\begin{minipage}[t]{10.5cm}
\small
\texttt{Wait, that seems contradictory because 16 is greater than 16.888.}\\
\texttt{That can't be. Wait, maybe I misapplied the bounds.}\\
\texttt{[...]}\\
\texttt{But in the initial manual addition, I had the sum as about 16.888,}\\
\texttt{but in reality, it's about 17.28. Therefore, the answer is 17.}
\end{minipage} \\
\midrule
\textbf{Note} & The student correctly computes $\sum \approx 16.888$ at one point, then contradicts this with an erroneous re-approximation, ultimately reporting $\lfloor 17.28 \rfloor = 17$ instead of $\lfloor 16.888 \rfloor = 16$. \\
\bottomrule
\end{tabular}
\end{table*}

\end{document}